\documentclass[letterpaper]{article} 
\usepackage{aaai2026}  
\usepackage{mathptmx}   
\usepackage{lmodern}    

\usepackage{lmodern}   

\usepackage{times}  
\usepackage{helvet}  
\usepackage{courier}  
\usepackage[hyphens]{url}  
\usepackage{graphicx} 
\usepackage{natbib}  
\usepackage{caption} 
\usepackage{algorithm}
\usepackage{algorithmic}
\newcommand{\zl}{\textcolor{black}}

\usepackage{amssymb}
\usepackage{amsmath}
\usepackage{amsthm}
\usepackage{booktabs} 

\newtheorem{definition}{Definition}
\newtheorem{theorem}{Theorem}

\newtheorem{remark}{Remark}

\usepackage{newfloat}
\usepackage{listings}
\DeclareCaptionStyle{ruled}{labelfont=normalfont,labelsep=colon,strut=off} 
\floatstyle{ruled}
\newfloat{listing}{tb}{lst}{}
\floatname{listing}{Listing}
\title{Fairness-Aware Design for Contextual Experiments: \\Guaranteeing Reliability and Equity in Heterogeneous Subgroups}
\author {
    Guangyan Gan \footnote{Corresponding author} \textsuperscript{\rm 1,\rm 2} ,
    Ling Zhang \textsuperscript{\rm 3}, Yanhua Cheng \textsuperscript{\rm 2},
    Yongxiang Tang \textsuperscript{\rm 2},
    Kaiyuan Li   \textsuperscript{\rm 2},\\
    Xialong Liu\textsuperscript{\rm 2},
    Peng Jiang  \textsuperscript{\rm 2}
}

\affiliations {
    \textsuperscript{\rm 1} School of Physical and Mathematical Sciences, Nanyang Technological Unviersity, Singapore\\
    \textsuperscript{\rm 2} Kuaishou Technology, China\\
     \textsuperscript{\rm 3}School of Management, Beijing Institute of Technology, China\\
    guangyan001@e.ntu.edu.sg, zhangling@bit.edu.cn, chengyanhua@kuaishou.com, tangyongxiang@kuaishou.com, likaiyuan03@kuaishou.com, liuxialong2007@sina.com,jp2006@139.com
}

\usepackage{bibentry}

\begin{document}

\maketitle

\begin{abstract}
Experimental design is critical for evidence-based decision-making in healthcare, marketing, and public policy. However, designing efficient experiments across heterogeneous subgroups presents significant challenges. Existing methods often optimize for statistical power or overall sample efficiency, overlooking crucial fairness considerations across these different subgroups. To address this gap, we introduce a Fairness-Aware Contextual Track-and-Stop Design (F-CTSD) algorithm. The proposed F-CTSD algorithm provides statistical guarantees on subgroup fairness while minimizing required sample sizes. We quantify the fairness-efficiency trade-off and derive the exact sample complexity for the proposed F-CTSD algorithm under its fairness constraints. We further theoretically prove that the proposed F-CTSD algorithm consistently produces accurate treatment effect estimates even under fairness requirements, enhancing statistical reliability. Numerical experiments show that the proposed F-CTSD algorithm outperforms existing methods, achieving higher sample efficiency while reducing subgroup fairness violations by 4.95\%.
\end{abstract}

\section{Introduction}
\label{sec:introduction}

Experimental design plays a pivotal role across various domains, including public policy \citep{bond201261, opper2019does, viviano2025policy}, healthcare \citep{chick2022bayesian, alban2023value}, and digital platforms \citep{johari2022experimental, bojinov2023design}. In these settings, experimenters often leverage data-driven algorithms to adaptively optimize interventions and improve operational efficiency. Adaptive allocation has proven to be more efficient than traditional random experiments, such as classical randomized controlled trials  \citep{lai1985asymptotically}. However, a substantial body of research in the social sciences documents pronounced \emph{heterogeneity} in treatment effects across different groups \citep{angrist2004treatment, varadhan2013estimation, wager2018estimation, kunzel2019metalearners, simchi2024experimentation}. This heterogeneity is particularly consequential in areas like personalized healthcare and recommendation systems, where the effectiveness of interventions varies significantly across subpopulations. As a result, strategies that tailor treatments based on observable characteristics have gained increasing attention.

Most existing methodologies in experimental design focus primarily on statistical objectives, such as maximizing power, minimizing variance, and reducing bias, often overlooking fairness considerations, especially when decisions affect heterogeneous subgroups \citep{simchi2024experimentation}. This neglect can systematically disadvantage certain groups, raising significant ethical and regulatory issues.
Documented real-world examples, 
such as discriminatory pricing in e-commerce \citep{cohen2025dynamic, xu2023doubly} and biased treatment allocations in healthcare \citep{chien2022multi}, highlight the importance of integrating fairness into experimental design to ensure equitable opportunities and outcomes.
Recent advances, such as the non-parametric causal forest estimator \citep{wager2018estimation}, enable robust estimation of heterogeneous treatment effects. Additionally, \citet{simchi2024experimentation} optimize experimental design to efficiently identify the best treatment per subgroup under minimal sample complexity, leveraging the Best Arm Identification (BAI) framework while explicitly modeling subgroup heterogeneity. Despite these strides, a critical challenge remains underexplored: fairness in experimental design, particularly in the context of heterogeneous and distinct subgroups.

Fairness concerns are paramount when allocation policies induce disparities across sensitive attributes such as age, gender, or race \citep{viviano2024fair}. In practical applications, including personalized pricing, clinical trials, and public policy, experimental designs that optimize solely for efficiency risk generating unfair treatment allocations across subgroups. 
These disparities not only undermine ethical standards but also raise concerns related to regulatory compliance and the credibility of experimental outcomes.
Incorporating fairness into experimental design is driven by both ethical imperatives and practical benefits. Ensuring equitable treatment allocation safeguards vulnerable populations and fulfills social responsibilities, while fairness-aware designs enhance the external validity and acceptance of experimental findings, particularly in high-stakes fields like healthcare and public policy. However, enforcing fairness constraints introduces inherent trade-offs: mitigating disparities often reduces statistical efficiency or restricts the adaptive allocation of participants to more effective treatments. This tension calls for principled methods that explicitly balance efficiency and fairness in treatment allocation.

In this paper, we address experimental design under explicit fairness constraints within the contextual BAI framework, focusing on a fixed-confidence setting. Our goal is to develop principled algorithms that identify the optimal treatment for each subgroup while rigorously enforcing fairness constraints throughout the learning process.

We present the following contributions:

\begin{enumerate}
    \item \textbf{Sample Complexity Lower Bound.} We derive an instance-dependent lower bound on the sample complexity for any $\delta$-weighted-Probably Approximately Correct (PAC) algorithm that satisfies fairness constraints. This bound quantifies the price of fairness, referring to the additional samples required to identify the optimal arm while adhering to fairness constraints, shading light on the trade-off between sample complexity and fairness.
    
    \item \textbf{Algorithm Design.} We propose the F-CTSD algorithm under the $\delta$-weighted PAC constraint. Our algorithm asymptotically matches the sample complexity lower bound and ensures that fairness constraints are satisfied at each interaction round for pre-specified fairness values. Unlike the Fair Best Arm Identification (F-BAI) algorithm \citep{russo2024fair}, we introduce a novel stopping rule tailored to the $\delta$-weighted PAC constraint. Additionally, our tracking procedure is probabilistic, contrasting with the deterministic methods used in the Contextual Track-and-Stop (CTSD) algorithm \citep{simchi2024experimentation}, which only identifies the optimal arm without fairness considerations.
    
    \item \textbf{Numerical Experiments.} We evaluate F-CTSD on synthetic datasets, showing that our algorithm outperforms benchmark methods in both sample efficiency and minimizing fairness violations.
\end{enumerate}

By rigorously incorporating fairness into experimental design, our work provides both methodological innovations and practical insights for conducting equitable and efficient experiments in heterogeneous populations.

\section{Related Work}
\label{sec:related}

Our work intersects three primary research areas: adaptive experimental design, best arm identification, and fairness. 
We provide an overview of each in Appendix.

\section{Fairness-Ware Experimental Design}
\label{sec:problem_setting}

This section formulates the adaptive experimental design problem using a contextual bandit framework and studies a fairness-ware experimental design problem.

\subsection{Adaptive Experimental Design}
We study an adaptive experimental design problem within a contextual bandit framework. At each discrete time step \( t \), a single experimental unit arrives with contextual information \( x_t \) from a discrete and finite contextual space \( \mathcal{X} := \{1, 2, \ldots, M\} \). Each \(x_t\) corresponds to a subgroup based on pre-treatment characteristics. For a given problem instance \( \boldsymbol{\nu} \), we assume \( x_t \) is independent and identically distributed (i.i.d.) generated from an unknown distribution \( \mathcal{P}_{\boldsymbol{\nu},x} \), where $\mathbb{P}_{\boldsymbol{\nu}}(x_t = x) = p_{\boldsymbol{\nu},x} > 0$ for all $x \in \mathcal{X}$.
Upon observing $x_t$, the experimenter assigns a treatment $a_t \in \mathcal{A} := \{1, 2, \ldots, K\}$. The environment generates potential outcomes $r_{x_t,a}$ for each treatment option $a \in \mathcal{A}$, but only reveals the reward $r_{x_t,a_t}$ associated with the selected treatment $a_t$. At time $t + 1$, the experimenter can only utilize the observed information collected in the history
\(
\mathcal{H}_t = (x_1, a_1, r_{x_1,a_1}, \ldots, x_t, a_t, r_{x_t,a_t}).
\)
This history forms a filtration $\{\mathcal{H}_t\}_{t \geq 1}$ updated per treatment.
For subgroup $x$ and treatment $a$, let $\nu_{x,a}$ denote the distribution of reward $r_{x,a}$ with mean $\theta_{x,a}$. We formally define the instance for subgroup $x$ as $\boldsymbol{\nu}_x = (\nu_{x,1}, \ldots, \nu_{x,K})$ and its mean reward vector as $\boldsymbol{\theta}_x = (\theta_{x,1}, \ldots, \theta_{x,K})^\top$. Therefore, every
problem instance can be formally denoted by $\boldsymbol{\nu} = (\mathcal{P}_{\boldsymbol{\nu},x}, (\boldsymbol{\nu}_x)_{x \in \mathcal{X}})$, and all the possible instances forming the set $\mathcal{S}$.


\zl{
We adopt the standard identifiability condition in adaptive experimental design \citep{simchi2024experimentation}: for each subgroup $x \in \mathcal{X}$, there exists a unique optimal treatment 
\(
a_x^*(\boldsymbol{\nu}) := \arg \max_{a \in \mathcal{A}} \theta_{x,a}.
\)
This heterogeneity condition reflects empirical evidence that optimal treatments systematically differ across subgroups \citep{obermeyer2016predicting, lada2019observational, imai2013estimating}.
The experimenter aims to output an optimal treatment \(\hat{a}_x\) for each subgroup \(x\) such that \( \hat{a}_x = a^*(\boldsymbol{\nu}_x) \) with high statistical confidence. Aligning with \citet{simchi2024experimentation}, we formalize this guarantee through a \(\delta\)-weighted Probably Approximately Correct (PAC) constraint. 
}

\zl{
\begin{definition}[\(\delta\)-weighted-PAC constraint \citep{simchi2024experimentation}]
\label{def: WPAC}
For any instance $\nu$ in $\mathcal{S}$, $\hat{a}_{x}$ satisfies
\begin{equation}
\label{eq:WPAC}
    \sum_{x \in \mathcal{X}} p_{x,\boldsymbol{\nu}}  \mathbb{P}_{\boldsymbol{\nu}}\left(\hat{a}_x \neq a^*(\boldsymbol{\nu}_x)\right) \leq \delta.
\end{equation} 
\end{definition}
 The parameter $\delta$ ensures that statistical guarantees hold with high confidence. This bound controls the decision error, where weighting by subgroup proportions \(p_{x,\boldsymbol{\nu}}\) imposes stricter accuracy requirements for larger subgroups.
To achieve this constraint, we minimize the expected experimental budget quantified by the stopping time \(\tau_\delta\), the terminal round where final decisions \(\hat{a}_x\) are implemented. Given unit arrivals per round, \(\tau_\delta\) equivalently gives the total sample size. 
}

\subsection{Fairness-Aware Experimental Design}

Adaptive experimental design problems with \(\delta\)-weighted-PAC constraint aim to output an optimal treatment for each subgroup with minimal sample complexity (i.e., minimizing the stopping time \(\tau_\delta\)). However, they overlook crucial fairness considerations across these subgroups. Widely deployed healthcare algorithms exhibit racial bias: Black patients assigned identical risk scores as white patients show greater illness severity due to cost-based proxies, making them 28.8\% less likely to qualify for additional care \citep{obermeyer2019dissecting}. This arises when experimental designs prioritize cost prediction over equitable outcomes

\zl{
To address systemic disparities, we formulate fairness constraints ensuring minimum selection rates for each treatment within every subgroup. These constraints prevent systematic under-selection and maintain equitable treatment allocation throughout the experiment. Let \( N_{x,a}(t) = \sum_{s=1}^t \mathbf{1}_{\{x_s = x, a_s = a\}} \) denote the number of times arm $a \in \mathcal{A}$ has been selected under subgroup $x \in \mathcal{X}$ up to time \( t \), and \( N_x(t) = \sum_{s=1}^t \mathbf{1}_{\{x_s = x\}} \) be the total number of observations with subgroup $x$. We denote the $q$-fairness constraints in experimental design problems.
}

\zl{
\begin{definition}[$q$-fairness constraints]
\label{def: fairness}
The $q$-fairness constraint requires that the expected allocation proportion of each arm $a \in \mathcal{A}$ within every subgroup $x \in \mathcal{X}$ exceeds a predetermined threshold $q_{x,a} \in [0,1]$, i.e.,
\begin{equation}
\label{eq:fairness1}
\frac{\mathbb{E}_{\boldsymbol{\nu}}[N_{x,a}(\tau_\delta)]}{\mathbb{E}_{\boldsymbol{\nu}}[N_x(\tau_\delta)]} \geq q_{x,a}, \quad \forall a \in \mathcal{A}, \quad x \in \mathcal{X}.
\end{equation} 
\end{definition}
}

\zl{
When \( q_{x,a} = 0 \), Eq.~\eqref{eq:fairness1} holds trivially, allowing optimal strategies
to select each subgroup's best treatment without fairness considerations. Conversely, \( q_{x,a} > 0 \) forces full treatment exploration, explicitly trading fairness for sample efficiency: Lower values weaken fairness guarantees, while higher values enforce stricter assignment equity (i.e., guaranteed minimum allocation proportions).
}

\zl{
Our fairness framework generalizes and strengthens prior work in three main aspects. First, whereas \citet{russo2024fair} introduce single-group \(\delta\)-PAC fairness constraints requiring minimum treatment probabilities, we explicitly incorporate subgroup structure and define fairness at the subgroup level. This enables modeling heterogeneous treatment effects while enforcing equity across subgroups, making our approach strictly more expressive for stratified settings. Second, while \citet{wei2024fair} address envy-freeness in adaptive designs, their focus on statistical power fundamentally neglects experimental cost control. Third, our formulation directly minimizes the expected sample size—quantified by the stopping time \(\tau_\delta\)—under joint fairness and \(\delta\)-weighted-PAC constraint.
}

\zl{
Integrating the $\delta$-weighted PAC constraint with $q$-fairness constraints, we define a $q$-fair $\delta$-weighted-PAC algorithm as follows:
\begin{definition}[$q$-fair $\delta$-weighted-PAC algorithm]
\label{def:fair_delta_weighted_PAC_algorithm}
An algorithm is $q$-fair $\delta$-weighted-PAC if for all $\boldsymbol{\nu} \in \mathcal{S}$ and $\delta \in (0,1/2)$,
it satisfies:
\begin{enumerate}
    \item [(1)] \textbf{$\delta$-weighted-PAC constraint} (Definition \ref{def: WPAC}),
    \item [(2)] \textbf{$q$-fairness constraints} (Definition \ref{def: fairness}),    
    and 
    \item [(3)] \textbf{finite stopping:} 
    \(
    \mathbb{P}_{\boldsymbol{\nu}}(\tau_\delta < \infty) = 1.
    \)
\end{enumerate}
\end{definition}
This definition formalizes fairness-aware experimental design. The fairness constraints prevent allocation disparities by ensuring minimum exposure $q_{x,a}$ for each arm-subgroup pair. The $\delta$-weighted-PAC constraint ensures statistical reliability of the selected treatments. That is, with probability at least \( 1 - \delta \), the algorithm correctly identifies the best treatment for each subgroup, with subgroup-level errors weighted by their importance \( p_{x,\boldsymbol{\nu}} \). The finite stopping condition ensures termination. Together, these yield efficient, equitable algorithms that generalize the $\delta$-weighted-PAC framework \cite{simchi2024experimentation} with subgroup fairness.
}


\zl{
The experimenter seeks a \( q \)-Fair \(\delta\)-weighted-PAC algorithm that minimizes the expected sample size \(\mathbb{E}_{\boldsymbol{\nu}}[\tau_\delta]\) while satisfying:
\[
\begin{array}{cl}
\min & \mathbb{E}_{\boldsymbol{\nu}}[\tau_\delta] \\
\text{s. t.} & \displaystyle \sum_{x \in \mathcal{X}} p_{x,\boldsymbol{\nu}} \mathbb{P}_{\boldsymbol{\nu}}(\hat{a}_x \neq a^*_x(\boldsymbol{\nu})) \leq \delta, \\
& \displaystyle \frac{\mathbb{E}_{\boldsymbol{\nu}}[N_{x,a}(\tau_\delta)]}{\mathbb{E}_{\boldsymbol{\nu}}[N_x(\tau_\delta)]} \geq q_{x,a}, \quad \forall a \in \mathcal{A},  x \in \mathcal{X}.
\end{array}
\]
The $\delta$-weighted-PAC constraint guarantees reliable identification of optimal treatments $a^*_x(\boldsymbol{\nu})$ across subgroups, while the $q$-fairness constraints enforce minimum allocation thresholds $q_{x,a}$ per subgroup-treatment pair---preventing systematic neglect of therapeutic options within any subgroup.
}


\section{\zl{Fairness-Aware Contextual Track-and-Stop Design Algorithm 
}}
\label{sec:algo}


\zl{
This section proposes a Fairness-Aware Contextual Track-and-Stop Design (F-CTSD) algorithm, extending recent advances in fairness-aware best arm identification \citep{russo2024fair} to contextual settings. Our framework integrates $q$-fairness constraints into contextual best arm identification under fixed-confidence regimes. We theoretically prove that the proposed F-CTSD algorithm is a $q$-Fair $\delta$-weighted-PAC algorithm, belonging to the Track-and-Stop (TaS) family \citep{garivier2016optimal}.
}


\zl{
The proposed F-CTSD algorithm comprises three core components: sampling rule, stopping rule, and selection rule.
We detail each component below and give the pseudo-code in the Appendix.
}

\subsection{Sampling Rule}
\label{sec:sampling_rule}

\zl{
We define the sampling rule \( \pi = \{\pi_t\}_{t \geq 1} \) as a policy 
mapping history \( \mathcal{H}_{t-1} \cup \{x_t\} \) to treatment \( a_t \), governing experimental unit allocation. The core insight is that sampling proportional to \( \omega_{x,q}^\star\left(\boldsymbol{\theta}_x\right) \) simultaneously minimizes sample complexity while satisfying fairness constraints. 
Since the mean reward vector \( \boldsymbol{\theta}_x \) is unknown, we implement this via:
\begin{enumerate}
    \item [(1)] \textbf{Parameter estimation}: Incrementally update reward estimates
    \[
    \hat{\theta}_{x, a}(t) := \frac{\sum_{s=1}^t r_{x_s, a_s} \cdot \mathbf{1}[x_s = x, a_s = a]}{N_{x, a}(t)}, 
    \]
    \[
    \boldsymbol{\hat{\theta}}_x(t) = \left( \hat{\theta}_{x,1}(t), \dots, \hat{\theta}_{x,K}(t) \right)^\top,
    \]
    with strong consistency \( \boldsymbol{\hat{\theta}}_x(t) \xrightarrow{\text{a.s.}} \boldsymbol{\theta}_x \) (Lemma 1 in Appendix).
    \item [(2)] \textbf{Instance mapping}: Convert to exponential family instance
    \[
    \boldsymbol{\hat{\nu}}_x(t) \leftarrow \boldsymbol{\hat{\theta}}_x(t) \quad \text{(via canonical parameterization)}.
    \] 
    \item [(3)] \textbf{Proportion optimization}: Solve the fairness-constrained lower bound by
    \[
    \boldsymbol{\omega}_{x,q}^\star(t) = \underset{\omega_x \in \Sigma_q}{\arg\max} \inf_{\nu'_x \in \operatorname{Alt}(\boldsymbol{\hat{\nu}}_x)} \sum_{a \in \mathcal{A}} \omega_{x,a} \, \mathrm{KL}(\hat{\nu}_{x,a}, \nu'_{x,a}) .
    \]   
    \item [(4)] \textbf{Best arm identification}: 
    \[
    a_{x,t}^\star = \arg \max_{a \in \mathcal{A}} \hat{\theta}_{x,a}(t).
    \]
\end{enumerate}
The sampling rule then allocates arms to track \( \boldsymbol{\omega}_{x,q}^\star(t) \) while respecting fairness constraints.
}

To enforce that the parametric uncertainty asymptotically goes to $0$  (i.e., \( \boldsymbol{\hat{\theta}}_{x}(t) \to \boldsymbol{\theta}_{x} \) almost surely), we employ a probabilistic tracking mechanism, inspired by the method proposed in \citet{russo2024fair}. Specifically, we mix the estimated optimal allocation \( \boldsymbol{\omega}_{x,q}^\star(t) \) with a fixed policy \( \pi_{x,c} \) via a decaying coefficient \( \epsilon_t = \frac{1}{2t^2} \), yielding:
\[
\pi_x(t) = (1 - \epsilon_t) \cdot \boldsymbol{\omega}_{x,q}^\star(t) + \epsilon_t \cdot \pi_{x,c}.
\]
The constant policy \( \pi_{x,c} \) guarantees every arm is sampled infinitely often and is defined by:
\[
\pi_{x,c,a} =
\begin{cases}
q_{x,a}, & \text{if } q_{x,a} > 0, \quad K_{x0} \neq 0, \\
\frac{1 - q_x^{\mathrm{sum}}}{K_{x0}}, & \text{if } q_{x,a} = 0, \quad K_{x0} \neq 0, \\
q_{x,a} + \frac{1 - q_x^{\mathrm{sum}}}{K}, & \text{if } K_{x0} = 0,
\end{cases}
\]
where \( K_{x0} = \left\vert \{a \in [K] : q_{x,a} = 0\} \right\vert \).
Unlike traditional TaS algorithms that rely on deterministic tracking \citep{garivier2016optimal, simchi2024experimentation}, our approach samples actions probabilistically from \( \pi_x(t) \). This probabilistic nature simplifies implementation and naturally aligns with fairness constraints, eliminating the need for external exploration mechanisms. Lemma  1 and Theorem 7 in Appendix establish the strong consistency of our parameter estimates.

\subsection{Stopping Rule}
\label{sec:stopping_rule}
In this subsection, we introduce a novel stopping rule specifically designed to align with the structure of the fairness-aware contextual best arm identification setting.
The stopping rule determines when sufficient statistical evidence has been gathered to satisfy the $q$-fair $\delta$-weighted-PAC constraint, thereby signaling the end of the experiment. Formally, we define a stopping time $\tau_\delta$ with respect to the filtration $\left\{ \mathcal{H}_t \right\}_{t \geq 1}$, which governs the evolution of observations over time.

We first define the set of alternative instances:
\[
\operatorname{Alt}(\boldsymbol{\nu}_x) := \left\{ \boldsymbol{\nu}' \in \mathcal{S}_x \,\middle|\, a^*(\boldsymbol{\nu}_x) \neq a^*(\boldsymbol{\nu}') \right\},
\]
which includes all instances with a different optimal arm.
We also define a complexity measure that captures the distinguishability of the current instance from alternative ones under fairness constraints:
\begin{equation}
\label{eq:Tq_star}
T_q^*(\boldsymbol{\nu}_x) := \max_{\boldsymbol{\omega}_x \in \Sigma_q} \inf_{\boldsymbol{\nu}_x' \in \operatorname{Alt}(\boldsymbol{\nu}_x)} \sum_{a \in \mathcal{A}} \omega_{x,a} \mathrm{KL}(\nu_{x,a}, \nu_{x,a}'),
\end{equation}
where
\[
\Sigma_q := \left\{ \boldsymbol{\omega}_x \in [0,1]^K \,\middle|\, \sum_{a=1}^K \omega_{x,a} = 1,\ \omega_{x,a} \ge q_{x,a},\ \forall a \in \mathcal{X} \right\},
\]
and \( \mathrm{KL}(\nu_{x,a}, \nu_{x,a}') \) denotes the Kullback-Leibler divergence between distributions \( \nu_{x,a} \) and \( \nu_{x,a}' \).
We also define, for any feasible allocation \( \boldsymbol{\omega}_x \in \Sigma_K := \left\{ \boldsymbol{\omega}_x \in [0,1]^K \mid \sum_{a=1}^K \omega_{x,a} = 1 \right\} \),
\[
T(\boldsymbol{\nu}_x, \boldsymbol{\omega}_x) := \inf_{\boldsymbol{\nu}_x' \in \operatorname{Alt}(\boldsymbol{\nu}_x)} \sum_{a \in \mathcal{A}} \omega_{x,a} \mathrm{KL}(\nu_{x,a}, \nu_{x,a}'),
\]
and denote by \( \boldsymbol{\omega}_x^*(\boldsymbol{\nu}_x) \) the optimizer of the maximization in \eqref{eq:Tq_star}, which satisfies
\[
T_q^*(\boldsymbol{\nu}_x) = T(\boldsymbol{\nu}_x, \boldsymbol{\omega}_x^*(\boldsymbol{\nu}_x)).
\]
At the population level, we define the overall complexity as the weighted sum over all groups:
\begin{equation}
\label{eq:Tq_global}
T_q^* := \sum_{x=1}^M p_{\boldsymbol{\nu}, x} T_q^*(\boldsymbol{\nu}_x),
\end{equation}
where \( p_{\boldsymbol{\nu}, x} \) is the probability of group \( x \) under the instance \( \boldsymbol{\nu} \).
We slightly abuse notation by using \( \boldsymbol{\omega}_x^*(\boldsymbol{\theta}_x) \) and \( \boldsymbol{\omega}_x^*(\boldsymbol{\nu}_x) \) interchangeably, as the one-parameter canonical exponential family induces a bijection between the natural parameter \( \boldsymbol{\theta}_x \) and the distribution \( \boldsymbol{\nu}_x \).

To ensure valid termination, any admissible stopping rule must satisfy the following weighted-PAC inequality $\forall x, \, \boldsymbol{\nu}_x^{\prime} \in \operatorname{Alt}\left(\boldsymbol{\nu}_x\right)$:
\[
\sum_{x=1}^M p_{\nu, x} \exp \left\{-\sum_{a=1}^K \mathbb{E}_\nu\left[N_{x, a}\left(\tau_\delta\right)\right] \mathrm{KL}\left(\boldsymbol{\nu}_x, \boldsymbol{\nu}_x^{\prime}\right)\right\} \leq 4 \delta.
\]
This expression motivates the use of an empirical analogue as a practical and implementable stopping criterion. 
Inspired from \cite{simchi2024experimentation}
we define the stopping time as the first round \( t \) such that:
\begin{equation}
\label{eq:stopping_rule}
\sum_{x=1}^M \hat{p}_x^{\prime}(t) \exp \left\{-\inf _{\nu_x^{\prime} \in \operatorname{Alt}\left(\hat{\nu}_x\right)} \sum_{a=1}^K N_{x, a}(t) \mathrm{KL}\left(\hat{\nu}_{x, a}, \nu_{x, a}^{\prime}\right)\right\} 
\end{equation}
$$\leq \phi(t, \delta),$$
where \( \hat{p}_x^{\prime}(t) := \frac{N_x(t)}{t} + 4 \sqrt{\frac{\log \log (t)}{t}} \) is an optimistic estimate of the true group weight \( p_{\nu, x} \), and \( \phi(t, \delta) = \frac{\delta}{t^{3K}} \) is a time- and confidence-dependent threshold function. We show that this rule ensures satisfaction of the $q$-fair $\delta$-weighted-PAC constraint in Theorem \ref{prop:p_delta_PAC}.

Importantly, our stopping rule operates jointly across all groups, rather than applying a separate rule to each individual group. This group-wise aggregation ensures that the stopping condition is met globally, thereby promoting statistical parity across contexts and preserving the algorithm's fairness guarantees at termination.

\subsection{Selection Rule}
\label{sec:selection rule}
The selection rule is a measurable mapping from the observed history at stopping time, $\mathcal{H}_{\tau_\delta}$, to the set of arms $\mathcal{A}^M$, yielding a final decision $\hat{a}_x$ for each context group $x \in \mathcal{X}$. Once the stopping criterion is satisfied, the F-CSTD algorithm proceeds by selecting the arm with the highest empirical mean reward for each context. Formally, for every \( x \in \mathcal{X} \), the selected arm is given by:
\begin{equation}
\label{eq:selection rule}
\hat{a}_x = \arg \max _{a \in \mathcal{A}} \hat{\theta}_{x, a}\left(\tau_\delta\right).
\end{equation}
This rule ensures that the algorithm outputs the empirically best arm at the stopping time, reflecting the accumulated evidence under the fairness-aware and context-sensitive sampling process.



 \section{Theoretical Analysis}
\label{sec:theoretical_analysis}

In this section, we first establish an instance-dependent lower bound on the sample complexity that holds for any \( q \)-fair \( \delta \)-weighted-PAC design. These bounds characterize the minimal number of samples necessary to ensure fair and probably approximately correct outcomes across heterogeneous groups. We then quantify the price of fairness in Section \ref{sec:price_of_fairness}, which reflects the increase in sample complexity induced by fairness constraints.
We next show that our proposed F-CSTD is \( q \)-fair \( \delta \)-weighted-PAC in Section \ref{sec:fairness}. Then, we establish the sample complexity guarantees of F-CSTD in Section \ref{sec:sample_complexity_guarantees}, and finally analyze the inference accuracy of treatment effects in Section \ref{sec:inference}.

\subsection{Sample Complexity Lower Bound}
\label{sec: Sample complexity lower bound}

The following theorem states a lower bound on the sample complexity of any $q$-fair $\delta$-weighted-PAC algorithm. The key quantity driving the lower bound is the \emph{characteristic time} \( T_q^* \) in \eqref{eq:Tq_global}. 

    
\begin{theorem}[Instance-dependent Lower Bound]
\label{thm:sample_complexity_lb}
Let \( \delta \in (0,1) \). For any \( q \)-fair \( \delta \)-weighted-PAC algorithm and any instance \( \boldsymbol{\nu} \in \mathcal{S} \), the expected stopping time satisfies:
\[
\mathbb{E}_{\boldsymbol{\nu}}[\tau_\delta] \ge \frac{\log \left(\frac{1}{4\delta}\right)}{T_q^*}.
\]
\end{theorem}

\begin{remark}
When \( q = (0,\ldots,0) \), the fairness constraint is inactive and the result recovers the lower bound of \cite{simchi2024experimentation}. In the special case where \( M = 1 \), Theorem~\ref{thm:sample_complexity_lb} reduces to the fairness-aware lower bound of \cite{russo2024fair}. For \( M = 1 \) and \( q = (0,\ldots,0) \), the result recovers the classical lower bound for Best-Arm Identification (see, e.g., \citet{garivier2016optimal,kaufmann2016complexity}).
\end{remark}

\subsection{The Price of Fairness}
\label{sec:price_of_fairness}
We now quantify the cost incurred by enforcing fairness constraints through the following result, which bounds the relative increase in sample complexity. Specifically, we upper-bound the ratio
\(\frac{T_{q_0}^*}{T_q^*}, \)
which measures the additional cost (in terms of sample complexity) of imposing fairness constraints compared to the unconstrained setting.
\begin{theorem}[Price of Fairness]
\label{thm:price_of_fairness}
Let \(  q = (q_{x,a})_{x \in \mathcal{X}, a \in \mathcal{X}} \in [0,1]^{M \times K} \) be a nontrivial fairness constraint, and define \( q_0 := (0,\ldots,0)^{M \times K} \) to denote the unconstrained case. Let \( q_{\min} := \min_{x,a} q_{x,a} \). Then,
\begin{equation}
\label{eq:price_fairness}
1 \le \frac{T_{q_0}^*}{T_q^*} \le \frac{1}{q_{\min}}.
\end{equation}
\end{theorem}

\begin{remark}
Theorem~\ref{thm:price_of_fairness} confirms that fairness constraints inevitably increase the sample complexity: for any \( q \ne q_0 \), it holds that \( T_q^* \le T_{q_0}^* \). This highlights a fundamental trade-off between fairness and efficiency.
Moreover, the upper bound \( \frac{1}{q_{\min}} \) implies that the price of fairness remains finite, even in the worst case. Stricter fairness requirements (i.e., larger values of \( q_{\min} \)) result in a larger increase in sample complexity, reflecting a higher cost for more equitable treatment.
\end{remark}

\subsection{Fairness Guarantees} 
\label{sec:fairness}
We first present the following main result, which establishes that F-CTSD satisfies the $q$-fair $\delta$-weighted-PAC property.
\begin{theorem}[$q$-fair $\delta$-weighted-PAC Constraint]
\label{prop:p_delta_PAC}
Let F-CTSD be run with \( \phi(t, \delta) = \frac{\delta}{t^{3K}} \). Then, F-CTSD is $q$-fair $\delta$-weighted-PAC. In particular:
\begin{itemize}
    \item \textbf{Fairness:} For all \( t \ge 1 \), \( x \in \mathcal{X} \), and \( a \in \mathcal{X} \),
    \[
    \frac{\mathbb{E}[N_{x,a}(t)]}{N_x(t)} \ge q_{x,a}.
    \]
    \item \textbf{$\delta$-weighted-PAC :} For every instance \( \boldsymbol{\nu} \in \mathcal{S} \),
    \[
    \sum_{x \in \mathcal{X}} p_{\boldsymbol{\nu}, x} \, \mathbb{P}_{\boldsymbol{\nu}} \left( \tau_\delta < \infty,\ \hat{a}_x \ne a^*(\boldsymbol{\nu}_x) \right) \le \delta.
    \]
\end{itemize}
\end{theorem}
Theorem~\ref{prop:p_delta_PAC} confirms that F-CTSD guarantees fairness at every round and satisfies the desired $\delta$-weighted-PAC  constraints.
\subsection{Sample Complexity Guarantees}
\label{sec:sample_complexity_guarantees}
We now analyze the efficiency of F-CTSD. Specifically, we show that it achieves the optimal sample complexity asymptotically as \( \delta \to 0 \), matching the lower bound in Theorem~\ref{thm:sample_complexity_lb}.
\begin{theorem}[Asymptotic Optimality in Expectation]
\label{thm:asymptotic_optimality}
Let F-CTSD be run with \( \phi(t, \delta) = \frac{\delta}{t^{3K}} \). Then, for all \( \delta \in (0, 1/2) \), the expected sample complexity is finite, i.e.,
\(\mathbb{E}_{\boldsymbol{\nu}}[\tau_\delta] < \infty. \)
Furthermore, for every instance \( \boldsymbol{\nu} \in \mathcal{S} \), the following asymptotic guarantee holds:
\(\lim_{\delta \to 0} \mathbb{E}_{\boldsymbol{\nu}}[\tau_\delta] \le \frac{\log \left( \frac{1}{4\delta} \right)}{T_q^*}. \)
\end{theorem}
Theorem~\ref{thm:asymptotic_optimality} shows that for sufficiently small \( \delta \),  the expected stopping time of F-CTSD closely matches the lower bound. This suggests that approximately $\frac{\log \left(\frac{1}{4 \delta}\right)}{T_{\boldsymbol{q}}^*}$ samples are nearly necessary and sufficient to ensure \( q \)-fair \( \delta \)-weighted-PAC in expectation. In other words, the expected duration of the experiment scales proportionally with \( \log(1/\delta) \), and the constant of proportionality is precisely governed by the inverse of \( T_q^* \), which encodes the problem’s inherent difficulty under fairness constraints. The detailed proof is deferred to the Appendix.

\begin{theorem}[Almost Sure Asymptotic Optimality]
\label{thm:almost_sure_optimality}
For any instance \( \boldsymbol{\nu} \in \mathcal{S} \), F-CTSD satisfies:
\[
\mathbb{P}_{\boldsymbol{\nu}}\left( \lim_{\delta \to 0} \tau_\delta \le \frac{\log \left( \frac{1}{4\delta} \right)}{T_q^*} \right) = 1.
\]
\end{theorem}
Theorem~\ref{thm:almost_sure_optimality} is stronger than Theorem~\ref{thm:asymptotic_optimality}, as it guarantees the same sample complexity bound almost surely, not merely in expectation. Specifically, it asserts that the stopping time \( \tau_\delta \) of F-CTSD satisfies
\(\tau_\delta \leq \frac{\log \left( \frac{1}{4\delta} \right)}{T_q^*} \)
with probability 1 as \( \delta \to 0 \). This result implies that F-CTSD achieves the optimal sample complexity almost surely in the asymptotic regime. Consequently, the algorithm not only performs well in expectation but also offers robust guarantees on individual realizations, even under fairness constraints. The detailed proof is provided in the Appendix.

\subsection{Inference}
\label{sec:inference}
A central objective in experimentation is conducting reliable statistical inference, particularly estimating the mean outcome of each treatment arm. Accurate mean estimation further enables valid inference on treatment effects. This section analyzes the concentration of the empirical mean \( \hat{\theta}_{x,a}(t) \) around the true mean \( \theta_{x,a} \), thereby establishing the inferential reliability of F-CTSD.
We show that F-CTSD guarantees high-probability confidence bounds for mean estimates under fairness constraints. The detailed proof is deferred to the Appendix.
\begin{theorem}[Mean Estimation Guarantee]
\label{thm:inference}
Let \( x \in \mathcal{X} \), \( a \in \mathcal{X} \), and \( \alpha \in (0,1) \). Then, with probability at least \( 1 - \alpha \),  the empirical mean \( \hat{\theta}_{x,a}(t) \) satisfies:
\[
\mathbb{P}\left(  d\left( \hat{\theta}_{x,a}(t), \theta_{x,a} \right) \le \frac{3 \log\left( \frac{\log(N_x(t) q_{x,a} + 1) + 1}{\alpha} \right)}{N_x(t) q_{x,a}} \right) \ge 1 - \alpha,
\]
where \( d(\cdot, \cdot) \) denotes a suitable divergence function, and \( N_x(t) \) denotes the number of samples collected under context \( x \) by time \( t \).
\end{theorem}
Theorem~\ref{thm:inference} implies that the confidence interval for the mean narrows as \( N_x(t) \) and \( q_{x,a} \) increase. We observe that a larger \( q_{x,a} \) (i.e., stricter fairness constraints) leads to a smaller error bound, and hence tighter confidence intervals. This highlights an additional benefit of fairness constraints: although they increase the overall sample complexity and hence the experimental cost, they also promote uniform exploration across arms. This uniformity results in more balanced data and, consequently, more accurate and reliable inference of treatment effects.
\section{Numerical Experiments}
\label{sec:NumericalExperiments}
\begin{figure}[t]
\centering
\includegraphics[width=1.0\columnwidth]{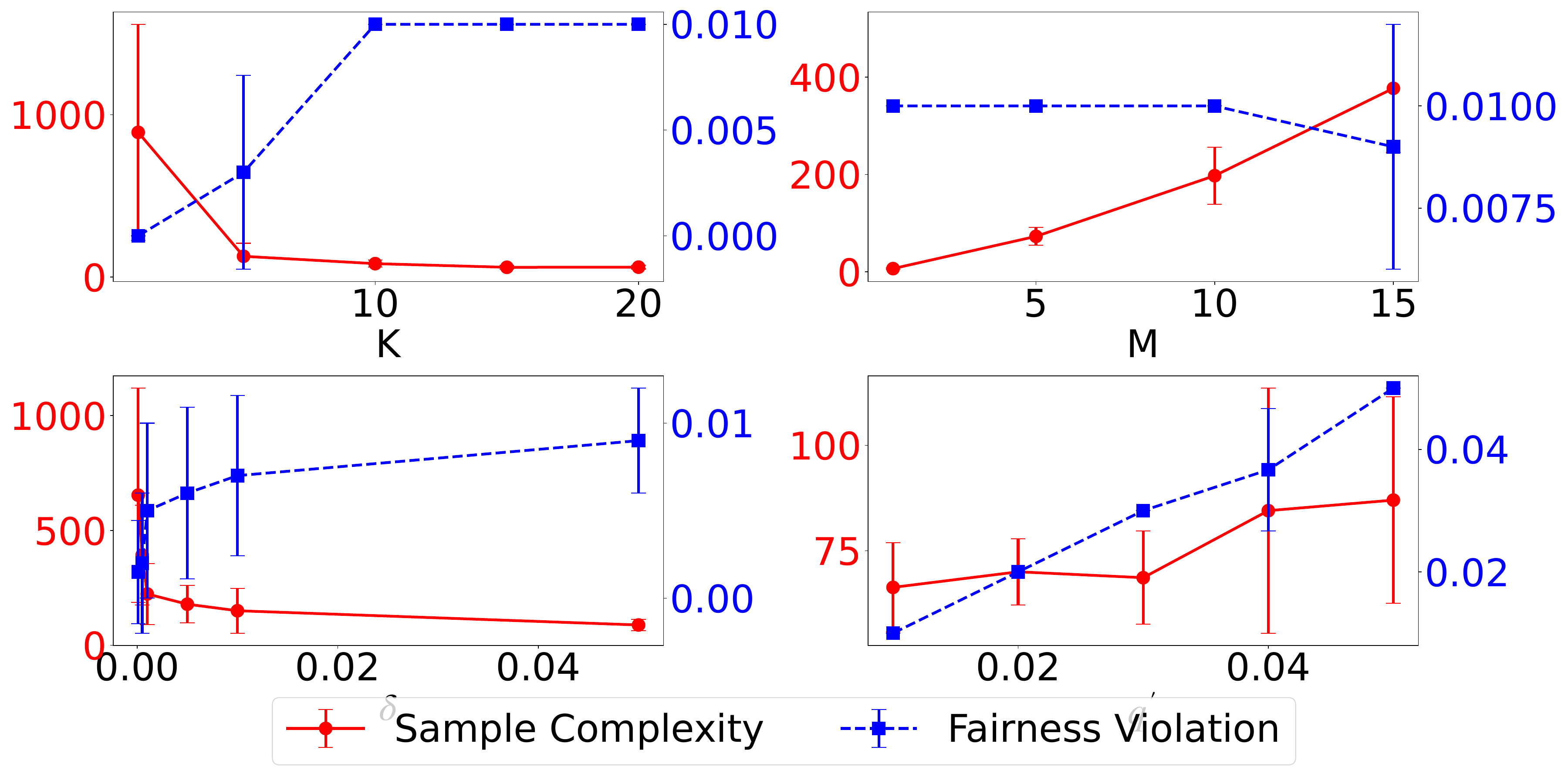} 
\caption{Impact of hyperparameters on synthetic data.}
\label{fig:fairness_violation_sample_complexity}
\end{figure}
In this section, we empirically evaluate the performance of our proposed F-CSTD algorithm on both synthetic and real-world datasets. The objective is to assess the algorithm’s effectiveness in balancing fairness and sample efficiency across diverse contexts.

\subsection{Experimental Setup}

\paragraph{Synthetic data.}  
We consider $K$ arms and $M$ contexts. The expected rewards $(\theta_{x,a})_{a \in \mathcal{X}}$ are linearly spaced in $[0,5]$, and the fairness vector is set as $q = q' [1,\dots,1]$, with $q' \in [0,1/K]$. We vary the number of contexts $M \in \{1, 5, 10, 15, 20\}$ and consider multiple confidence levels $\delta \in \{10^{-5}, 10^{-4}, 10^{-3}, 10^{-2}, 10^{-1}\}$. Each configuration is repeated over $N = 50$ independent runs.

\paragraph{Real-world data.}  
We use the COVID-19 Clinical Trials dataset~\citep{larson2022covid}. Contexts are defined by study Gender, Age, and Phase, resulting in $M=29$ unique contexts. Interventions are encoded into $K=11$ categories. The reward is whether study results are posted (1) or not (0). The sequential learning setup mirrors that of synthetic experiments, with real-time updates of estimated success probabilities and fairness constraints.

\subsection{Baseline Algorithms}
We compare F-CSTD against three baselines:
\begin{itemize}
    \item \textbf{Contextual Track-and-Stop Design (CTSD)}~\citep{simchi2024experimentation}: Adapts to subgroup heterogeneity but does not enforce fairness.
    \item \textbf{Fair Best-Arm Identification (F-BAI)}~\citep{russo2024fair}: Enforces fairness at the population level, assuming a single context ($M=1$).
    \item \textbf{UniformFair}: Allocates sampling probability uniformly while satisfying minimal fairness thresholds: 
    \[
        \pi_{x,a}(t) = q_{x,a} + \frac{1 - q^{\mathrm{sum}}_x}{K}, \quad q^{\mathrm{sum}}_x = \sum_{a=1}^K q_{x,a}.
    \]This design guarantees that the expected sampling frequency satisfies  
\( \mathbb{E}_\nu\left[\frac{N_{x,a}(t)}{N_x(t)}\right] \geq q_{x,a} \quad \text{for all } t \geq 1.\)
\end{itemize}

\subsection{Evaluation Metrics}

We evaluate algorithms along two dimensions:

\begin{enumerate}
    \item \textbf{Sample Complexity:} The expected number of samples $\mathbb{E}_\nu[\tau_\delta]$ required to meet the confidence threshold $\delta$.
     \item \textbf{Fairness Violation:} Defined at stopping time $\tau_\delta$ as:
    $$
    \text{Fairness Violation} = \mathbb{E}_\nu\left[\rho\left(\tau_\delta\right)\right], \quad 
    $$
where $\rho(t) = \max\left(\max_{x,a} \left\{ q_{x,a} - \frac{N_{x,a}(t)}{N_x(t)} \right\}, 0 \right).$ 
    This measures the average maximum deviation from the prescribed allocation $q_{x,a}$ across all groups at the stopping time $\tau_\delta$.
\end{enumerate}

\begin{table}[htbp]
\centering

\begin{tabular}{lcc}
\toprule
Algorithm & Fairness Violation & Sample Complexity \\
\midrule
CTSD           & 0.0081 & 114.81 \\
F-BAI          & 0.0083 & 34.09  \\
UniformFair    & 0.0047 & 178.50 \\
\textbf{F-CSTD} & \textbf{0.0048} & \textbf{169.66} \\
\bottomrule
\end{tabular}
\caption{Performance Comparison on Synthetic Data ($q'=0.01$, $\delta=0.05$, $K=M=10$).}
\label{tab:synthetic_results}
\end{table}

\subsection{Results on Synthetic Data}
Table~\ref{tab:synthetic_results} summarizes the empirical comparison of fairness violation and sample complexity on synthetic data.
Our proposed F-CSTD algorithm maintains a strong balance between fairness and efficiency. It achieves a fairness violation of 0.0048, nearly matching the best baseline (UniformFair: 0.0047), while reducing sample complexity from 178.50 to 169.66. Compared to CSTD, F-CSTD reduces fairness violation by over 40\% with only a moderate increase in sample complexity, demonstrating the benefit of fairness-aware adaptive learning.
Although F-BAI yields the lowest sample complexity (34.09), it exhibits the highest fairness violation (0.0083), exceeding F-CSTD by more than 70\%. This suggests that fairness enforced only at the population level may fail to protect subgroups in heterogeneous environments.

We further analyze the impact of hyperparameters in Figure~\ref{fig:fairness_violation_sample_complexity}. Fairness violation increases with the number of arms $K$, with higher confidence levels $\delta$, and with looser fairness constraints. Conversely, it decreases as the number of contexts increases. For sample complexity, it decreases with larger $K$ and larger $\delta$, but increases with more contexts and stricter fairness constraints.


\subsection{Results on COVID-19 Clinical Trials Data}

\paragraph{Comparison with benchmarks}
We compared FCTSD with several benchmarks: CTSD, FBAI, UniformFair. Table~\ref{tab:real_results} reports the mean  fairness violations and sample complexity across 50 repeated experiments. 
FCTSD achieved an better tradeoff between fairness and sample efficiency.

\begin{figure}[t]
\centering
\includegraphics[width=1.0\columnwidth]{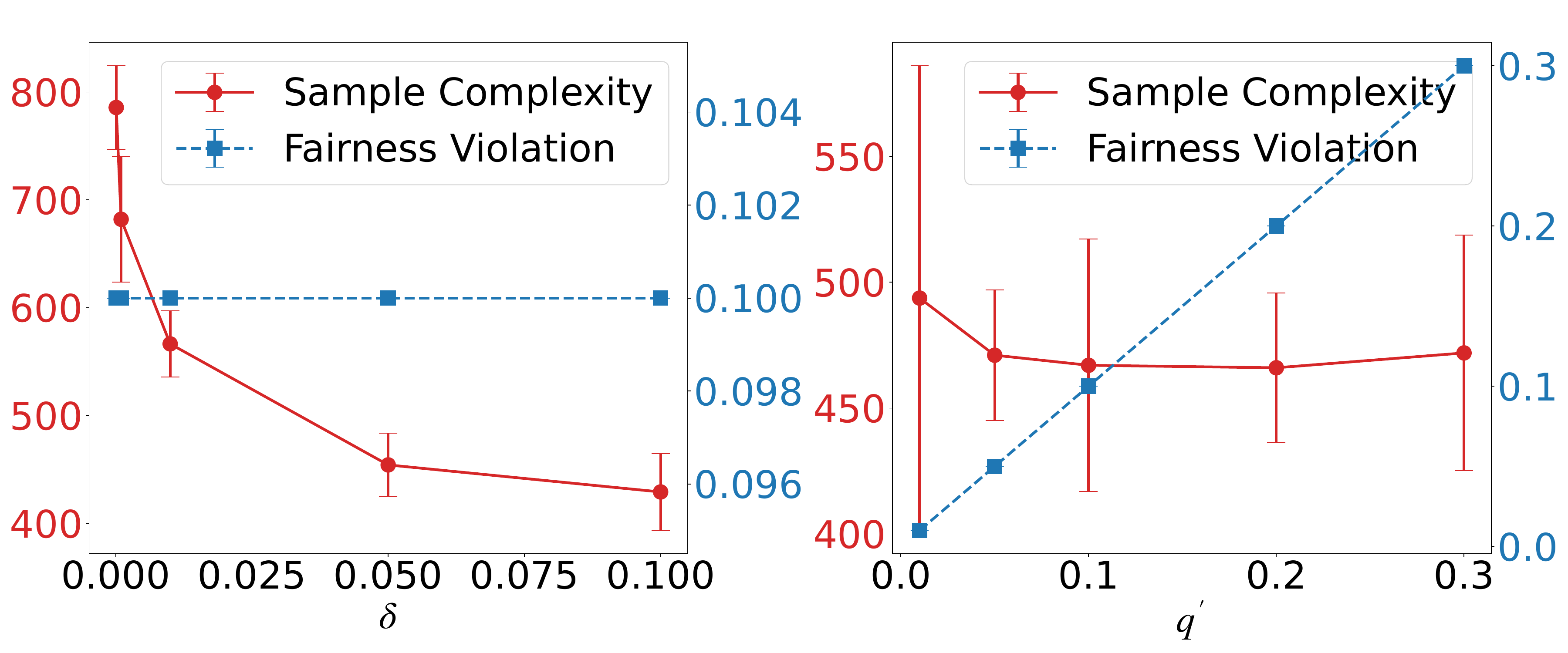} 
\caption{Impact of hyperparameters on clinical trials data..}
\label{fig:fairness_violation_sample_complexity_real}
\end{figure}

\paragraph{Sensitivity analysis.}
We performed a sensitivity analysis on key algorithmic parameters, including the fairness constraint threshold $q^\prime$ and the confidence parameter $\delta$. In Figure \ref{fig:fairness_violation_sample_complexity_real}, the trends observed in real-world data were consistent with those obtained from synthetic datasets, demonstrating that the algorithm's behavior is robust to parameter variations.

\begin{table}[htbp]
\centering

\begin{tabular}{lcc}
\toprule
Algorithm      & Fairness Violation & Sample Complexity \\
\midrule
CTSD            & 0.0124 & 256.73 \\
F-BAI           & 0.0118 & 102.44 \\
UniformFair     & 0.0079 & 341.55 \\
\textbf{F-CSTD}  & \textbf{0.0082} & \textbf{301.19} \\
\bottomrule
\end{tabular}
\caption{Performance Comparison on Real Data ($q'=0.01$, $\delta=0.05$, $K=11$, $M=29$).}
\label{tab:real_results}
\end{table}

\subsubsection{Discussion.}
F-CSTD consistently balances fairness and sample efficiency in both synthetic and real-world datasets. Sensitivity analyses confirm that algorithmic behavior is robust to variations in key hyperparameters ($q'$ and $\delta$). These results demonstrate the practical applicability of F-CSTD for adaptive decision-making scenarios where fairness across heterogeneous contexts is critical.

\section{Conclusion}
\label{sec: Conclusion}

In this work, we address the challenge of fair experimental design for heterogeneous subgroups by introducing the F-CTSD algorithm, which enforces explicit fairness constraints in treatment allocation. Our theoretical analysis establishes instance-specific lower bounds on sample complexity and quantifies the trade-off between fairness and efficiency. We further develop the F-CTSD algorithm, which provably achieves these lower bounds while ensuring fairness constraints are met throughout the experimental process. Extensive experiments on synthetic data confirm that our approach delivers a strong balance between fairness and sample efficiency, outperforming existing methods on both fronts.

While our framework advances the integration of fairness into adaptive experimental design, it primarily focuses on statistical objectives related to treatment identification. Future research could explore further aligning experimental objectives with the welfare and preferences of participants, as well as extending the framework to more complex or real-world settings.

\appendix
\section{Acknowledgments}
The authors gratefully acknowledge Hanzhang Qin and Zhenzhen Yan for their valuable comments and insights, as well as the fruitful discussions that greatly contributed to this work.
\bibliography{aaai2026}

\end{document}